\documentclass{optica-article}

\journal{opticajournal} 

\articletype{Research Article}

\usepackage{lineno}

\begin{document}

\title{Detecting immune cells with label-free two-photon autofluorescence and deep learning}


\author{Lucas Kreiss\authormark{1,2,3*}, Amey Chaware\authormark{1}, Maryam Roohian\authormark{4}, Sarah Lemire\authormark{3,5}, Oana-Maria Thoma\authormark{3,5}, Birgitta Carlé\authormark{2}, Maximilian Waldner\authormark{3,5}, Sebastian Schürmann\authormark{2}, Oliver Friedrich\authormark{2} and Roarke Horstmeyer\authormark{1}}

\address{\authormark{1}Computational Optics Lab, Department of Biomedical Engineering, Duke University, Durham, USA\\
\authormark{2}Institute of Medical Biotechnology, Department Chemical and Biological Engineering, Friedrich-Alexander University (FAU), Erlangen-Nürnberg, Germany\\
\authormark{3}Department Medicine I, University Hospital Erlangen, Germany \\
\authormark{4}Department of Pathology, Duke University, Durham, USA \\
\authormark{5}German Center for Immunotherapy, Deutsches Zentrum Immuntherapie (DZI), University Hospital Erlangen, Germany}

\email{\authormark{*}\href{mailto:lucas.kreiss@duke.edu}{lucas.kreiss@duke.edu} or 
\href{mailto:lucas.kreiss@fau.de}{lucas.kreiss@fau.de}} 


\begin{abstract*} 
Label-free imaging has gained broad interest because of its potential to omit elaborate staining procedures which is especially relevant for \textit{in vivo} use. Label-free multiphoton microscopy (MPM), for instance, exploits two-photon excitation of natural autofluorescence (AF) from native, metabolic proteins, making it ideal for \textit{in vivo} endomicroscopy. Deep learning (DL) models have been widely used in other optical imaging technologies to predict specific target annotations and thereby digitally augment the specificity of these label-free images. However, this computational specificity has only rarely been implemented for MPM. In this work, we used a data set of label-free MPM images from a series of different immune cell types (5,075 individual cells for binary classification in mixed samples and 3,424 cells for a multi-class classification task) and trained a convolutional neural network (CNN) to classify cell types based on this label-free AF as input. A low-complexity squeezeNet architecture was able to achieve reliable immune cell classification results (0.89 ROC-AUC, 0.95 PR-AUC, for binary classification in mixed samples; 0.689~F1 score, 0.697~precision, 0.748~recall, and 0.683~MCC for six-class classification in isolated samples). Perturbation tests confirmed that the model is not confused by extracellular environment and that both input AF channels (NADH and FAD) are about equally important to the classification. In the future, such predictive DL models could directly detect specific immune cells in unstained images and thus, computationally improve the specificity of label-free MPM which would have great potential for \textit{in vivo} endomicroscopy.

\end{abstract*}

\section{Introduction}

Inflammation is a key response of biological organisms to harmful stimuli, like pathogens or tissue damage. This mechanism is layered across different scales, ranging from macroscopic swelling of the 3D tissue down to the release of sub-cellular biochemical molecules. At the microscopic level, inflammation is governed by the recruitment and activation of a range of different immune cells. This process is highly dynamic and involves a specific timing of different immune cell types that are present at different locations, at different times, and in different amounts. The initial response during acute inflammation caused by tissue damage or by an unknown pathogen is typically governed by a series of innate immune cells, while inflammation through a specific pathogen might be identified and targeted by a chain of adaptive immune cells. Anomalous adaptive immune responses can lead to chronic inflammation and are often the root of many autoimmune diseases.

Therefore, the detection and classification of immune cells is essential to monitor inflammation dynamics, to enable a deeper understanding of the spatial organization of immune reactions or to diagnose autoimmune diseases and monitor the effect of potential treatment options.

Ideally, such detection tools should provide microscopic, cellular resolution and minimize alterations to the target cells, i.e., it is desirable to avoid fixation and sectioning as well as biochemical binding with antibody markers. Optical microscopy has developed several label-free imaging technologies that provide contrast from the natural interaction between biochemical structures and light. Many of those techniques have already been used for detection of specific cell types, often by leveraging the predictive power of deep learning (DL) models to boost specificity.

The most accessible and straightforward label-free technique is probably bright-field imaging (BF), which was used as input to DL models to classify type and state of certain cells with a performance comparable to fluorescence-based approaches~\cite{eulenberg2017reconstructing,harrison2023evaluating,yang2023live}. Phase contrast microscopy is another commonly used imaging technique where DL models have shown accurate results, for instance for the segmentation of bovine aortic endothelial cells~\cite{su2013cell}, classification of myoblast cells~\cite{niioka2018classification} or the classification of cancer cells~\cite{kang2025cancer}. Differential interference contrast (DIC) imaging provides enhanced contrast at edges and structural features by exploiting optical path length differences, producing pseudo-3D relief images of unstained cells~\cite{lee2017cell}. Deep learning has enabled classification and segmentation of cells~\cite{kuijper2008automatic}, their health status~\cite{pan2024accurate} or bacteria~\cite{obara2013bacterial} from DIC images. Ogawa \textit{et al.} compared the effect of different imaging modalities on DL-based classification between lymphoid-primed multipotential progenitor (LMPP) and pro-B cells~\cite{ogawa2022different}. They found no significant difference between BF, phase contrast, and DIC in a generally good classification performance (area under the receiver operating characteristic curve - AU-ROC of~0.9)~\cite{ogawa2022different}. Quantitative phase imaging (QPI) is a more advanced computational imaging technique that can provide intrinsic quantification of the optical path length difference to provide a quantitative imaging signal with decent cellular specificity~\cite{curl2005refractive,wang2010quantitative}. QPI has been used in a wide range of applications of machine learning-assisted cell assessment~\cite{park2023artificial,nguyen2022quantitative}, like classification of red blood cell morphology~\cite{jiang2022automatic}, scoring~\cite{lam2020quantitative} or classification of cancer cells~\cite{roitshtain2017quantitative}, distinction between healthy B cells and lymphoblasts~\cite{ayyappan2020identification}, as well as classification of stages in B cell acute lymphoblastic leukemia~\cite{ayyappan2020identification}. Similar advances have also been made with label-free digital holographic microscopy~\cite{wittenzellner2025label,go2018label,jaferzadeh2023automated}.

Despite their success in live cell microscopy, the translation of BF, phase imaging, DIC or QPI towards live tissue, 3D \textit{endo-}microscopy is less straight-foward, as these techniques either show limited 3D capabilities (BF and phase) or require multiple acquisitions under controlled illumination patterns (DIC and QPI), which is more challenging in a tightly-packed endoscope design, as well as computational reconstructions (DIC and QPI), which can be more error-prone for noisy \textit{in vivo} data. Furthermore, it is often desirable to obtain functional or metabolic information, especially for immune reactions. And although quantitative measurements of optical path length in QPI can be related to dry cell mass~\cite{kandel2020phase}, these image contrast quantities are only indirectly related to metabolic activities.

\begin{figure}
\centering
\includegraphics[width=\linewidth]{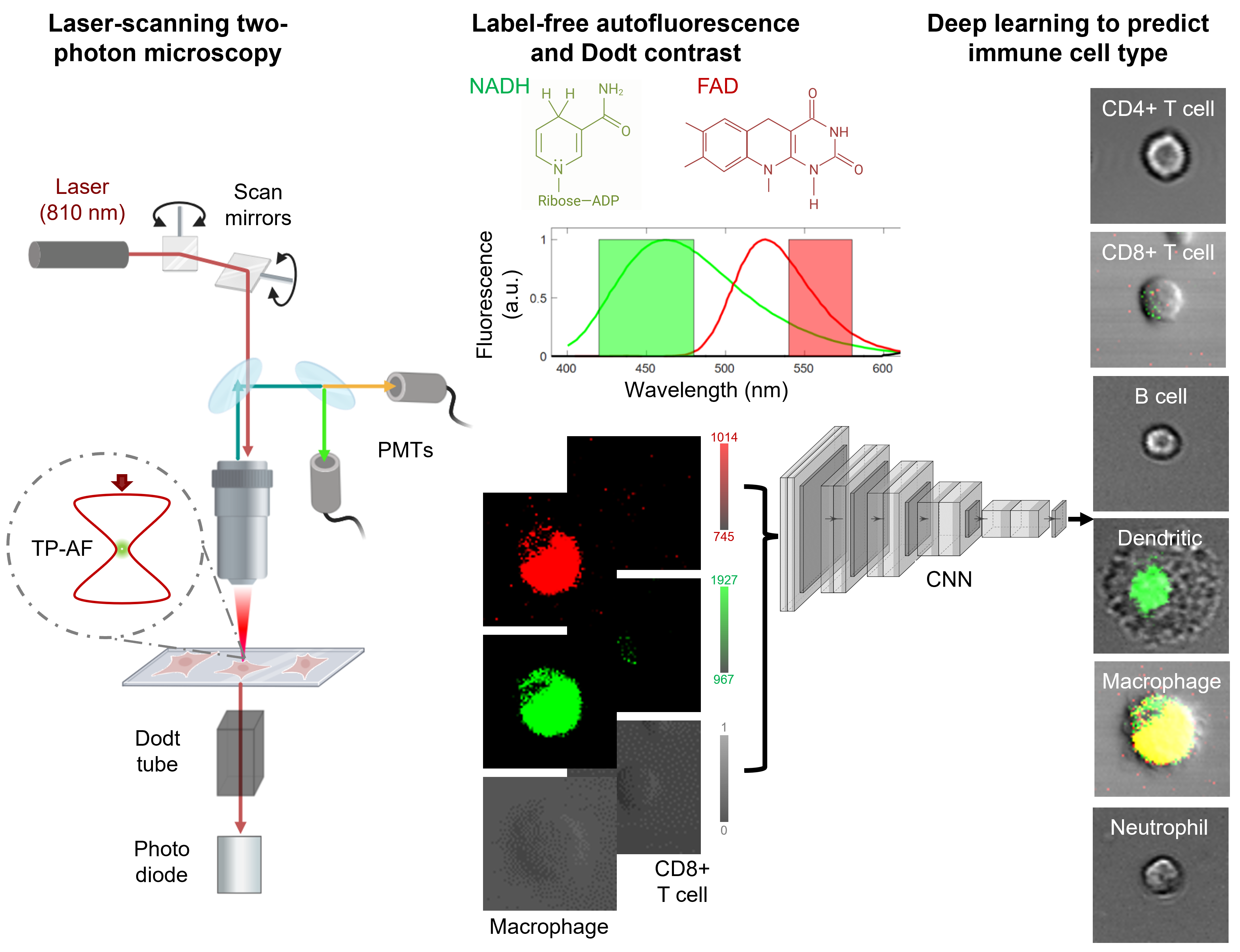}
\caption{\textbf{Automated immune cell identification based on label-free 2-Photon autofluorescence and deep learning.} A scanning multiphoton microscope is used to generate label-free image data from immune cells on a substrate. A 810~nm, ultra-short-pulsed laser is used to excite autofluorescence from NADH and FAD, while gradient Dodt contrast images are collected in transmission mode. Label-free images from various immune cells are collected and used as input to a convolutional neural network, which has been trained to predict immune cell type.}
\label{fig1}
\end{figure} 

In contrast, multiphoton microscopy (MPM)  exploits the confocal nature of nonlinear excitation for optical sectioning as well as reduced scattering at infra-red wavelength for greater penetration depths~\cite{zipfel2003nonlinear}. Moreover, the label-free measurement of natural, autofluorescence from metabolic coenzymes, like nicotinamide adenine dinucleotide (NADH, H for hydrogen) and flavin adenine dinucleotide (FAD), can directly be linked to metabolic processes and mitochondrial activity in cells~\cite{yu2009two}. These quantitative measurements have already been used to reveal cellular metabolic states~\cite{shi2025optical} or mitochondrial dysfunction~\cite{shi2017label}, to distinguish breast cancer cells from normal controls~\cite{yu2009two}, to identify different breast cancer tissues~\cite{shi2025optical}, to detect Alzheimer’s disease in fresh murine brain samples~\cite{shi2017label} or to support the distinction between brown and white adipose tissue~\cite{shi2018optical}. 

This unique combination of deep-tissue, 3D imaging capability paired with label-free, metabolic contrast, make MPM ideal for the investigation of inflammation and inflammatory tissue remodeling directly within the native 3D tissue structure~\cite{kreiss2020label,kreiss2022label,schmalzl2022interferon,sommer2023anti}. 

Although single-photon induced AF was already used for classification of immune cell types~\cite{habibalahi2025multispectral,shah2023autofluorescence}, cell sorting via flow cytometry~\cite{Lemire2022,abir2024metabolic} or digital staining of tissue sections and live cells~\cite{zhang2020digital,Rivenson_VS,kreiss2023digital}, two-photon induced AF has only rarely been explored for the same purposes. For instance, Gehlsen et al.~\cite{gehlsen2015non}, as well as our own group~\cite{Lemire2022} demonstrated distinction of various immune cell types by using statistical tools to distinguish two-photon AF intensities. However, the full potential of two-photon AF for AI-assisted computational specificity, as proposed for many other imaging modalities~\cite{zhang2020digital,Rivenson_VS,kreiss2023digital}, is still underexplored.
Label-free multimodal imaging of coherent-anti-stokes Raman scattering (CARS) and MPM
was used for digital H\&E staining based on the label-free input~\cite{bocklitz2016pseudo,pradhan2021computational}. However, these procedures were limited to formalin-fixed and paraffin-embedded tissue sections and did not offer specificity to different immune cell types beyond that of H\&E. The development of state-of-the-art DL models for specific immune cell classification based on label-free multiphoton images has not yet been demonstrated. 

Such computational specificity for immune cells in label-free multiphoton imaging might be particularly promising, since multiphoton imaging is already being used for \textit{in vivo} imaging and endo-microscopy (MPEM)~\cite{myaing2006fiber,bao2008fast,llewellyn2008minimally,rivera2011compact,huland2012vivo,dilipkumar2019label,kreiss2020label}, which enables optical histology imaging in live animal models. Thus, the development of AI-assisted immune cell detection for label-free multiphoton imaging might translate very well to \textit{in vivo} applications.

In this work, we used an existing data set of label-free, two-photon induced AF images from various different immune cell types~\cite{Lemire2022} to train CNN models on automated, specific identification of immune cells. The use of simultaneously recorded fluorescence antibody markers, as well as clear experimental designs allowed us to obtain reliable ground truth annotation without the common bottleneck of manual expert annotations. Systematic perturbation tests validated robust classification performance and data-efficient learning, while also revealing that models that focus on molecular two-photon induced AF significantly outperformed those that only use spatial information on cell size and shape alone.  

\section{Materials and Methods}

\subsection{Data set}
\label{section_data}

These original images were obtained in a previous work by Lemire et al.~\cite{Lemire2022} from immune cells that were isolated from the spleen (CD4$^+$/CD8$^+$ T cells, B cells) or bone marrow (macrophages, dendritic cells, neutrophils) of wildtype C57BL/6 mice, seeded on glass slides and imaged with a multiphoton microscope (TriMScope II, LaVision BioTec, Bielefeld, Germany) using filters that target AF from NADH (BP 450/70) and FAD (BP 560/40). The respective spectra and filter bands are shown in Fig.~\ref{fig1}a. In addition to these AF channels, the forward scattered Dodt channel was recorded (displayed in gray in all figures). Similar to DIC, Dodt contrast is a gradient-based technique, and it provides optical sectioning and improved visualization of thick tissue slices~\cite{dodt1990visualizing}. Although it offers enhanced structural detail, its application for DL-based immune cell classification is less established, and, as with DIC, it primarily encodes morphological rather than biochemical or metabolic differences. Each full raw image had a size of 1,024~$\times$~1,024 pixels across a field of view (FOV) of 405~$\times$~405~$\mu m^2$, containing dozens to hundreds of cells. Details of all data in this study are shown in table~\ref{tab_data}.

\paragraph{Cell mixture}
First, we investigated the potential to differentiate two different cell types (neutrophils and T cells) that were present in the same sample (Fig.~\ref{fig2}). In that case, T cells were stained with the allophycocyanin (APC)-labeled lymphocyte marker $\alpha$-CD3 to obtain ground truth annotations. This APC signal had no significant spectral overlap with natural AF emissions (BP 675/67 for APC, see Fig.~\ref{fig1}a) and was subsequently recorded at a different excitation wavelength (810~nm for AF and 1,040~nm for APC) which prevented channel leakage entirely. 

In total, this data set consisted of 31 full field-of-view image pairs of label-free AF images. Four images only contained T cells, seven images only contained neutrophils and 20 images were from samples that contained roughly a 50:50 mixture of both cell types. 

An image processing procedure was developed in the open-source image processing software Fiji to crop single cell image patches from these full-FOV images. The image processing macro loaded the raw data and registered AF and APC images via Scale Invariant Feature Transform (SIFT)~\cite{lowe2004distinctive}. Cell detection was performed via semi-automated, user-validated Otsu-thresholding of the NADH channel (`setAutoThreshold("Otsu no-reset")'), auto adjusted to include 10\% of bright pixels, and a human observer verified or adjusted the threshold manually, if needed. Thresholding was followed by Watershed and `Analyze Particles' (minimal size of 25~pixels area and 0.3~-~1 circularity). The center of the detected regions of interest (ROI) was then used to crop a patch of 64~$\times$~64 pixels (25.6~$\times$~25.6~$\mu m^2$ FOV) around it. For each image patch, the two AF channels and the Dodt channel were saved together as multi-channel TIF file. Cells at the edges of the original image were ignored to ensure a consistent size of 64~$\times$~64 for all patches. The respective APC channel of the patch was used for thresholding to determine it as APC-positive or APC-negative. This procedure resulted in a data set with a total of 5,078 annotated image patches, each with a unique cell in the center.

\paragraph{Multi-class data set}
In the second case, we investigated the potential of two-photon induced AF for multi-class classification of several different cell types. For that purpose, we used a different experimental design in the available data base~\cite{Lemire2022}, where cell types were not mixed, and each isolated cell type was imaged separately without antibody reference (Fig.~\ref{fig3}). Therefore, annotations for each cell type  were available from the experimental protocol instead of a fluorescence antibody (see Fig.~\ref{fig3}). Purity of these isolated cell suspensions reached values of >95\% in each case~\cite{Lemire2022}. In total, we pooled 85 unique full FOV images from six different cell types. These images were processed into 64~$\times$~64 pixel patches following the same procedure as explained above, resulting in a total of 3,424 cell patches.

\begin{table}[!htb]
    \caption{Data sets}
      \small
      \centering
        \begin{tabular}{lrr}
        \hline
        \textbf{} & \textbf{Binary classification} &  \textbf{Multi-class classification} \\
        
        \hline
         Full FOV images (1,064~$\times$~1,064 pixels)     & 31                         & 85 \\

         Image patches per class & CD4+ T cells: n=1,311      & CD4+ T cells: n=144 \\
         (64~$\times$~64 pixels) &                            & CD8+ T cells: n=612 \\
                                &                           & B cells: n=1,953 \\
                                & Neutrophils: n=3,764    & Neutrophils: n=366 \\
                                &                       & Dendritic cells: n=140 \\
                                &                       &  Macrophages: n=209 \\
        \hline
        Total image patches (64~$\times$~64 pixels)       & 5,075                       & 3,424 \\
    \end{tabular}
    
    \label{tab_data}
\end{table}

\subsection{Model architecture}
We selected the SqueezeNet architecture as the backbone model for this study, as it is known to preserve competitive accuracy while reducing the number of trainable parameters. In the case of our relatively small data set (see section~\ref{section_data}), this might minimize the risk of overfitting~\cite{hastie2009elements}. We used a pretrained Squeezenet architecture (squeezenet1.0, torchivision) and adjusted input, features and classifier layers (see Block 0, 1 and 11 in table~\ref{tab_model}) to match the shape of our input data and prediction labels. SqueezeNet is comprised mainly of `Fire' modules which are essentially squeezed convolution layers that only have 1x1 filters, feeding into an expand layer that has a mix of 1x1 and 3x3 convolution filters (for more details, see section 5.3 in Ref.~\cite{iandola2016exploring}). This pretrained model was then fine-tuned with our data set for the two respective classification tasks.

\subsection{Training}
The entire framework for training was developed in Python 3.9 using PyTorch 2.5.1 and CUDA 11.8. A custom data loader was used to load the TIF files for each cell patch as pytorch tensors, apply data augmentation (horizontal flip, vertical flip and rotation), apply a Gaussian filter ($\sigma$ = 2) and finally, to carry out a z-score standardization to the mean and standard deviation of the entire data set. A 5-fold cross validation (CV) was used, where 80\% of the data were used for training and the remaining 20\% for validation in each fold (sklearn, model selection, \textit{Kfold}). In case of the multi-class data, a grouped stratification (sklearn, model selection, \textit{stratifiedgroupkfold}) was used since this data set was severely imbalanced.

The models were trained for 300 epochs using a cross entropy loss, an Adam optimizer~\cite{kingma2017adammethodstochasticoptimization}, a learning rate scheduler and stochasic weight averaging (SWA)~\cite{izmailov2019averagingweightsleadswider}. As shown in supplementary Fig.~1, this resulted in a loss convergence. These main hyper parameters are summarized in table~\ref{tab_hyperparam}. In each fold, loss, true positives, true negatives, false positives and false negatives were tracked to calculate per-fold performance metrics which were averaged to the final evaluation.

\subsection{Data perturbation experiments}
In order to evaluate if the model learns the desired cellular information instead of overfitting to noise or extra-cellular background, we introduced a series of spatial perturbations in the binary classification data. Similarly to the approach by Cook et al.~\cite{cooke2022multiple}, all of these perturbations were in place for the entire network training process, resulting in an independently trained model for each perturbation test. As a spatial perturbation, we defined concentric circles with fixed diameters that masked either the inside or the surrounding area outside of that circle. The circles were always centered within the image and had diameters of 5, 20, 40, and 60~pixels (see Fig.~\ref{fig3}a). 

\subsection{Model perturbation experiments}
Our selected SqueezeNet architecture had a total of 735,937 trainable parameters. To evaluate whether this capacity was adequate for our tasks, we performed a model perturbation test, where certain blocks of the architecture were frozen to the initially pre-trained condition, without re-optimization through backpropagation on the new data ('requires\_grad = false'). The first input layer and the classifier block (blocks 0 and 11 in table~\ref{tab_model}) were always allowed for backpropagation, resulting in a minimum of 15,760 trainable parameters. We then increased the trainable capacity by subsequently 'un-freezing' one layer at a time, starting at \#3 in table~\ref{tab_model} (28,176 parameters) until \#10 (735,937 parameters).

\subsection{Channel perturbation experiments}
Finally, we evaluated the relative importance of the different input channels (representing different types of molecular/metabolic information) by training the same model architecture on four different input channel configurations: (i) NADH autofluorescence only, (ii) FAD autofluorescence only, (iii) Dodt gradient contrast only, and (iv) NADH and FAD autofluorescence. Again, an entirely new model was trained on each of these configurations.

\subsection{Computational Hardware}
All models were trained on a workstation equipped with NVMe SSD, Nvidia RTX 3090 GPU and Intel Core-i9 10850k CPU (10 cores of 3.6 GHz). Training took about 25~min~(1,500~s) for each of the perturbation experiments.

\section{Results}

\subsection{Classification of immune cells (in mixed samples)}

\begin{figure}
\centering
\includegraphics[width=\linewidth]{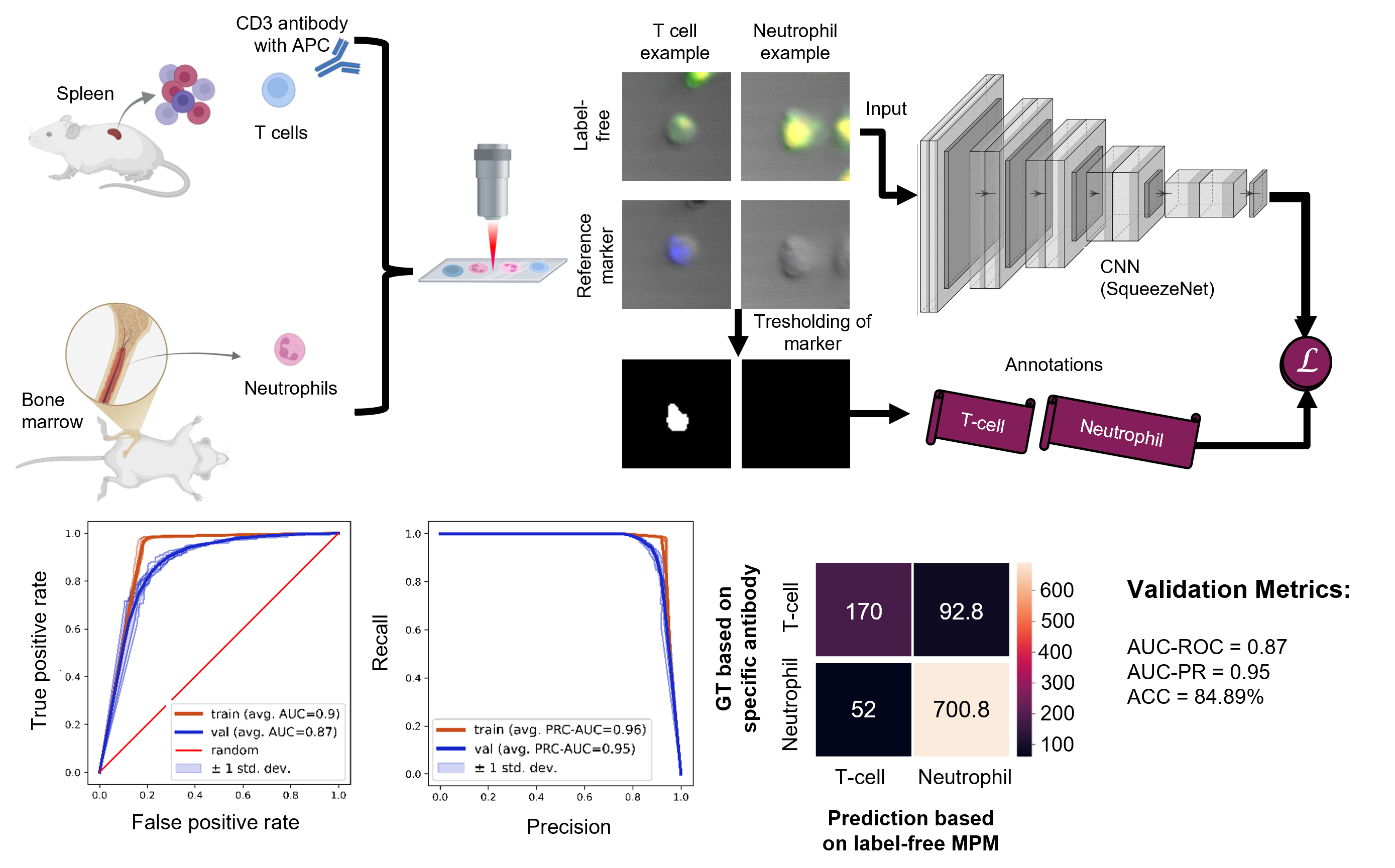}
\caption{\textbf{Classification results from unstained neutrophils and stained T cells in mixture.} T cells were isolated, stained with an APC-labeled $\alpha$-CD3 marker and mixed with unstained neutrophils, before imaging. The two label-free channels of NADH and FAD were used as input to a deep learning model, while the fluorescence channel of the specific marker was used to derive ground truth annotations for the training. Receiver operating characteristic (ROC) curve, precision-recall (PR) curve, and confusion matrix for the validation examples indicate that the model is able to differentiate both cell types reasonably well, when using label-free MPM images as input. All values denote the average across 5 folds. }
\label{fig3}
\end{figure}   

Fig.~\ref{fig3} shows the main classification results of the trained network. For the binary classification between T cells and neutrophils in mixed samples, we can report a 5-fold average AUC-ROC of 0.87, an AUC of the precision recall curve of 0.95 and a validation accuracy of 84.89\%. 

\subsection{Multi-class classification of other types of immune cells}

\begin{figure}
\centering
\includegraphics[width=\linewidth]{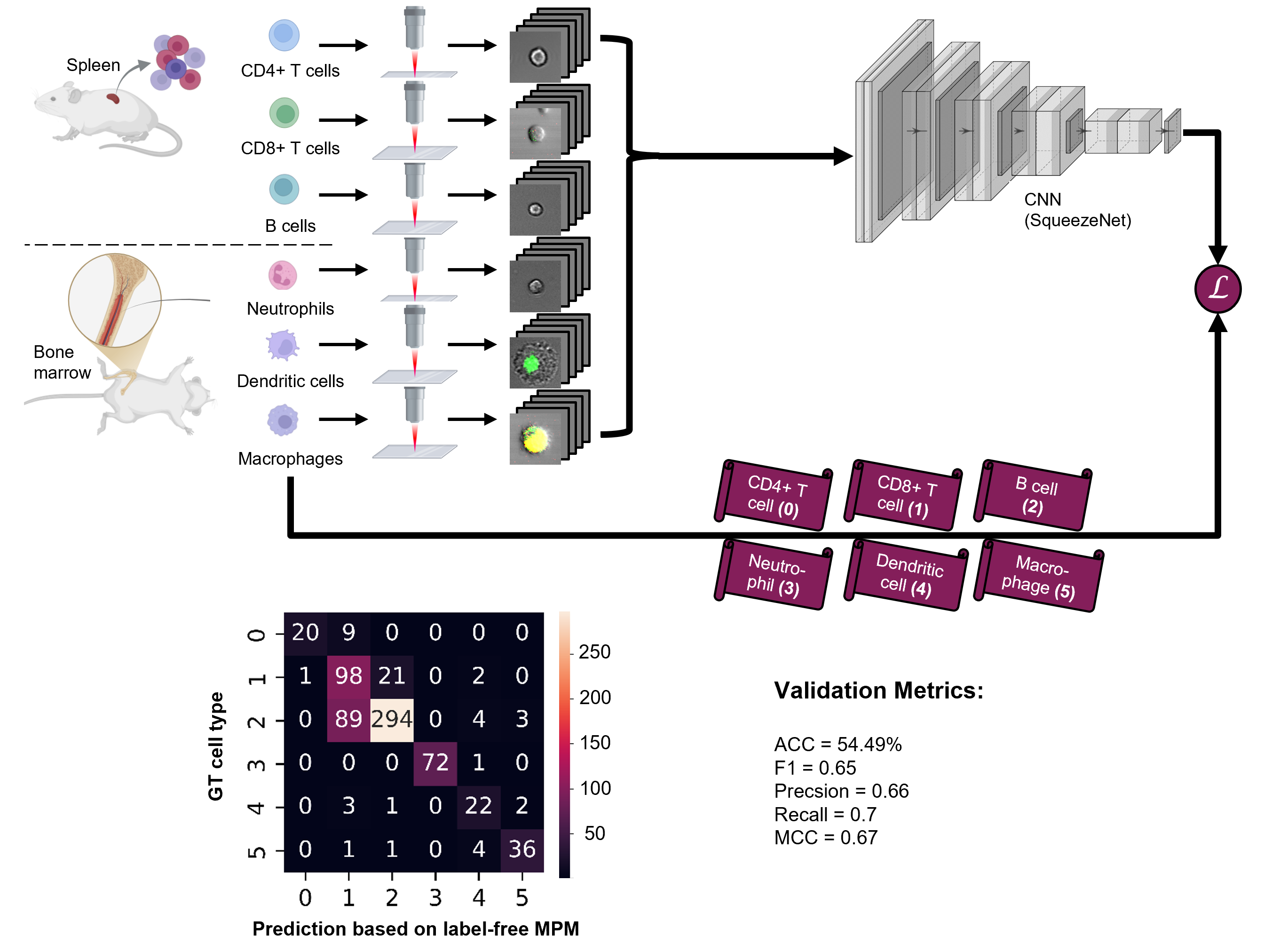}
\caption{\textbf{Classification results from six different immune cells.} Each cell type was isolated and imaged separately, resulting in a separate data set for each cell type, so that ground truth annotations were available through that experimental design. Again, a deep CNN model was trained with label-free AF images as input to predict cell type. Multi-class classification results are evaluated by the 5-fold cross-validation confusion matrix and performance metrics. All values denote the average across 5 folds. }
\label{fig2}
\end{figure} 

The multi-class cell classification shown in Fig.~\ref{fig2} generally shows successful classification results. The model achieved an F1 score of 0.689 (ranges from 0 to 1), a precision of 0.697 (range from 0 to 1), a recall of 0.748 (range from 0 to 1) and a Matthew's correlation coefficient (MCC) of 0.683 (range from -1 to 1). The multi-class accuracy was 52.67\% (random guess would be 16.6\%), and most of the mis-classified examples were B cells that were falsely predicted as CD8$^+$ T cells. As seen in Fig.~\ref{fig2} and supplementary Fig.~1, this multi-class data set was very skewed, with B cells being by far the largest class. Therefore, the metrics of F1 score, precision, recall and MCC are more conclusive metrics than the accuracy in this case. 

The limited size of the data set and the greater number of labels lead to the challenge of splitting examples across the five validation folds. The same K-fold CV strategy that was used for the binary classification (see above), would have resulted in folds that lacked representation of each class, leading to per-fold performance metrics, that are likely underestimating a potential performance on larger data. This is a known challenge, when splitting imbalanced data sets of multiple classes and few examples. There, we employed a grouped stratification strategy for this multi-class problem. Although this procedure reduced the reported validation performance, we prioritized methodological rigor by enforcing strict group-aware splitting to prevent data leakage - a conservative choice that sacrifices short-term metric optimization for long-term generalizability in this class-imbalanced setting.

\subsection{Perturbation experiments}

\begin{figure}
\centering
\includegraphics[width=0.75\linewidth]{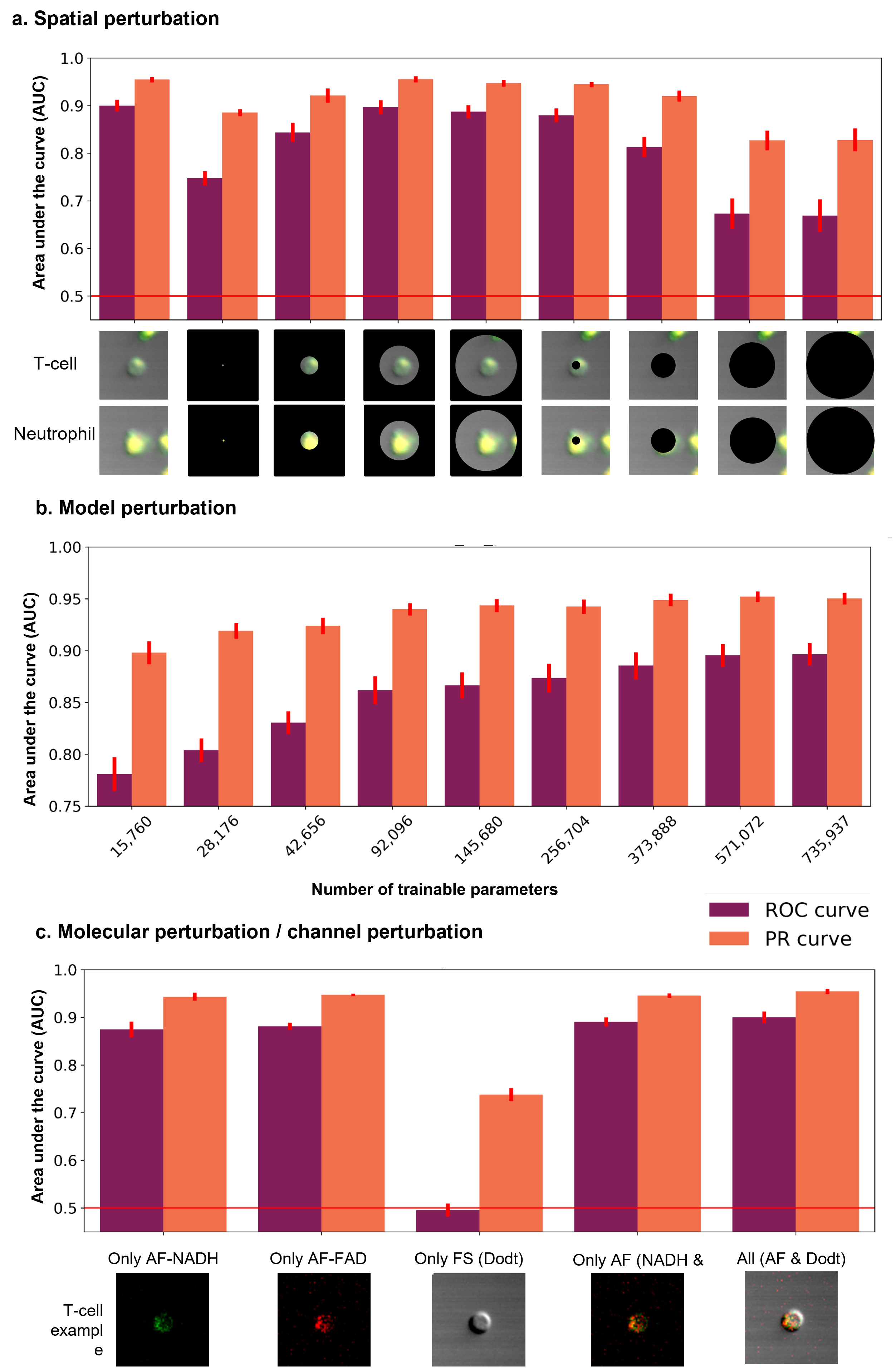}
\caption{\textbf{Perturbation experiments indicate that the model is learning the desired cellular autofluorescence pattern.} \textbf{(a)} Circles of different diameters were used to mask center or edge pixels. Examples of perturbation data are shown below for both cell types. \textbf{(b)} Performance changes with varying numbers of trainable parameters in the model (see table~\ref{tab_model}). \textbf{(c)} Performance changes for different molecular imaging channels. All performance values show area under the curve (AUC) for receiver operating characteristic (ROC) on the left and for precision-recall (PR) curve on the right from the validation case in a 5-fold cross validation scheme. Bar height represents the average validation performance of all folds, with the standard deviation as error bar.}
\label{fig4}

\end{figure}   

\paragraph{Data perturbation}

When presented with data that contain only a small fraction of central pixels (second bar in Fig.~\ref{fig4}a), the model performance drops from 0.87~AUC-ROC and a 0.95~PR-AUC (Fig.~\ref{fig4} \& the first bar in Fig.~\ref{fig4}a) to only 0.75~AUC-ROC / 0.89~PR-AUC. When trained on data with continuously larger circular areas from the center, the performance continuously increases, as expected, approaching a performance similar to that in the unperturbed case.
On the other hand, when trained on data in which central pixels were continuously \textit{removed}, the performance drops sharply to only 0.67~AUC-ROC/0.83~AUC-PR and slowly increases when allowing for more pixels. 
Together, these results indicate that learning heavily relies on the actual cellular information in the center of the image patch, as intended, and is not severely influenced by background pixels which contain noise and might occasionally include other cells at the edges.  


\paragraph{Model perturbation}
The results of our model perturbation tests, shown in Fig.~\ref{fig4}b, indicate that both ROC-AUC and PR-AUC steadily increase when increasing the number of trainable parameters. Moreover, this steady increase flattens out and converges at higher capacities of trainable parameters (i.e., between 571,072 and 735,937 trainable parameters), which indicates that the chosen model has an adequate capacity for the given task, and that a larger model is not expected to perform significantly better.

\paragraph{Cell detection is driven by cellular autofluorescence, not by shape}
Finally, we present the results of training the model on different image channels in Fig.~\ref{fig4}c. It can be observed that the performance of a model trained on only one single AF channel (i.e., NADH or FAD, respectively) is already close to the overall performance including all channels. A model that was solely trained on the Dodt signal is close to a random guess (0.5~AUC-ROC).


\section{Discussion}

This study was designed to develop deep neural networks for automated identification of specific immune cell types, based on label-free 2-photon induced AF image data. In the binary classification of mixed T-cells and neutrophils, we achieved a performance of 0.87~AUC-ROC and 0.95~AUC-PR which is on par with a recent state-of-the-art method for single-photon induced AF (0.92~-~1~AUC-ROC~\cite{habibalahi2025multispectral}), despite the fact that we only used three input channels instead of 56~\cite{habibalahi2025multispectral} and relied on a tiny data set of only 5,075 single immune cells. The second case of classifying six different major immune cell types also performed reasonably well (0.689~F1 score, 0.697~precision, 0.748~recall, and 0.683~MCC), especially since this task was a more complicated multi-class problem and since even fewer and yet more imbalanced data were available.

In addition to demonstrating an overall successful use of label-free MPM and DL for immune cell identification, our systematic perturbation tests enabled further investigation of robustness of the classification, as well as of the relative importance of different spatial-molecular patterns. The combination of controlled experiments and architectural choices ensured the model learned generalizable patterns rather than over-fitting to artifacts or background noise in the limited dataset. Moreover, molecular contrast perturbations indicated that learning was significantly more successful if the model had access to the full spatial distribution of molecular two-photon induced AF instead of only the spatial information from the gradient Dodt channel. 

The use of elegant data collection strategies allowed us to obtain high-quality data labels without human annotation and thus, to bypass manual annotations which can often be the labor-intensive bottle-neck in developing, training and validating new DL models. The use of reference fluorescence markers in combination with AF poses the unique challenge to avoid spectral overlap to preserve the specificity of the imaging signals and prevent information leakage. The presented approach to use subsequent recording of the APC channel at a different excitation can be used for two-photon excitation without overlapping with the natural AF emission. In the future, the use of genetically encoded reporter fluorophores would be a promising addition to provide cell annotations for the development of multiphoton imaging with computational specificity to other cells, like bacteria (e.g., \textit{Citrobacter rodentium}~\cite{schmalzl2022interferon}) or a finer resolution to different subtypes of immune cells (e.g., Ccr2$^+$/RFP, Cd68$^+$/GFP and Cx3cr1 macrophages~\cite{sommer2023anti}). The use of such genetically encoded fluorophores for digital staining has already been shown in the example of digital staining of mitochondria in living cells using correlative imaging~\cite{Somani2022} and would be most suitable for \textit{in vivo} imaging of immune cells in 3D tissues.

Following the good scientific practices suggested in Ref.~\cite{kreiss2023digital}, we aim to discuss the overall uncertainty of our approach which involves the fundamental labeling uncertainty in obtaining the ground truth data annotations and the prediction uncertainty of the model. The latter can be gauged by the standard deviation in performance across the different folds which is around 1-3\% (0.025~ROC-AUC or 0.013~PR-AUC). The label uncertainty however, is more difficult to assess. In the multi-class classification experiment, data labels were obtained through the experimental protocol of isolating the respective cells. It has been stated that the purity of these isolated cell samples reached >95\%~\cite{Lemire2022} which can be regarded as the upper boundary for the labeling specificity for this experiment. In the case of binary classification of T cells and neutrophils, the labeling specificity is related to the biochemical binding specificity of the antibody which is much more difficult to assess quantitatively. It is widely accepted that T cells carry CD3 antigens that bind with anti-CD3 antibodies, while neutrophils, B cells and macrophages do not. However, these statements are usually more qualitative, like "B cells, granulocytic series, and monocytes/macrophages are all CD3-negative."~\cite{NAEIM201829} or "the vast majority of mature T cells bear TCR $\alpha \beta$"~\cite{NAEIM201829}. Quantitative values for the half-maximal effective concentration (EC50) values of certain CD3 antibodies are available in literature to range from 3 to 24~nM~\cite{haber2021generation} or 17 to 161~nM~\cite{haber2021generation}. However, the actual specificity in percentage (\%), i.e., the percentage of selective binding to T cells against unintended binding to neutrophils, was not determined in this experimental study. Nevertheless, uncertainties in the data labeling is also an extremely challenging problem to conventional, manual annotations which are often subject to human errors, individual biases, or inter-observer variability~\cite{buckchash2025applications}.

\section{Conclusion \& Outlook}
In summary, we demonstrated the successful use of label-free MPM and DL for immune cell identification by training a CNN on label-free 2P-AF image data while obtaining reliable ground truth labels from the experimental design. Our simple yet effective model achieved good performance for differentiating T cells and neutrophils (0.87~AUC-ROC and 0.95~AUC-PR), as well as for multi-class classification of CD4$^+$ T cells, CD8$^+$ T cells, B cells, neutrophils, dendritic cells and macrophages (0.689~F1 score, 0.697~precision, 0.748~recall, 0.683~MCC). A series of different perturbation experiments shows that the model is stable (i.e., not confused by extracellular environment), efficient (i.e., has an adequate number of trainable parameters and learns from very small data sets), and that the detection is driven by cellular autofluorescence (NADH and FAD) instead of gradient Dodt contrast.

Computational specificity to immune cells, as proposed here, would be a promising addition to label-free multiphoton imaging. In the future, this concept could be applied to 3D deep-tissue images, for instance, to enable new fundamental research on inflammatory tissue remodeling. Moreover, the combination of this computational specificity with label-free \textit{in vivo} imaging via endomicroscopy would have great potential for clinical translation of this emerging technology.

\section{Back matter}

\begin{backmatter}
\bmsection{Funding}
This project has received funding from the European Union's Horizon 2022 Marie Sk\l{}odowska-Curie Action under grant agreement 101103200 (to LK), as well as from the German Research Foundation (DFG) under project C01 of the collaborative research center TRR-241 (to SS \& MW). 


\bmsection{Disclosures}
The authors declare no conflicts of interest.

\bmsection{Data Availability Statement}
The original data presented in the study will be made openly available in a Zenodo repository and the respective code will be made publicly available via Github, upon acceptance of this manuscript for publication.


\newpage

\end{backmatter}

\newpage

\section{References}
\label{sec:refs}

\bibliography{sample}

\begin{thebibliography}{10}
\newcommand{\enquote}[1]{``#1''}

\bibitem{eulenberg2017reconstructing}
P.~Eulenberg, N.~K{\"o}hler, T.~Blasi, \emph{et~al.}, \enquote{Reconstructing cell cycle and disease progression using deep learning,} {\protect\JournalTitle{Nature communications}} \textbf{8}, 463 (2017).

\bibitem{harrison2023evaluating}
P.~J. Harrison, A.~Gupta, J.~Rietdijk, \emph{et~al.}, \enquote{Evaluating the utility of brightfield image data for mechanism of action prediction,} {\protect\JournalTitle{PLOS Computational Biology}} \textbf{19}, e1011323 (2023).

\bibitem{yang2023live}
X.~Yang, D.~Chen, Q.~Sun, \emph{et~al.}, \enquote{A live-cell image-based machine learning strategy for reducing variability in psc differentiation systems,} {\protect\JournalTitle{Cell discovery}} \textbf{9}, 53 (2023).

\bibitem{su2013cell}
H.~Su, Z.~Yin, S.~Huh, and T.~Kanade, \enquote{Cell segmentation in phase contrast microscopy images via semi-supervised classification over optics-related features,} {\protect\JournalTitle{Medical image analysis}} \textbf{17}, 746--765 (2013).

\bibitem{niioka2018classification}
H.~Niioka, S.~Asatani, A.~Yoshimura, \emph{et~al.}, \enquote{Classification of c2c12 cells at differentiation by convolutional neural network of deep learning using phase contrast images,} {\protect\JournalTitle{Human cell}} \textbf{31}, 87--93 (2018).

\bibitem{kang2025cancer}
M.~Kang and J.~Kim, \enquote{Cancer cell classification based on morphological features of 3d phase contrast microscopy using deep neural network,} {\protect\JournalTitle{IEEE Access}}  (2025).

\bibitem{lee2017cell}
J.~Lee, I.~Kolb, C.~R. Forest, and C.~J. Rozell, \enquote{Cell membrane tracking in living brain tissue using differential interference contrast microscopy,} {\protect\JournalTitle{IEEE Transactions on Image Processing}} \textbf{27}, 1847--1861 (2017).

\bibitem{kuijper2008automatic}
A.~Kuijper and B.~Heise, \enquote{An automatic cell segmentation method for differential interference contrast microscopy,} in \emph{2008 19th International Conference on Pattern Recognition,}  (IEEE, 2008), pp. 1--4.

\bibitem{pan2024accurate}
F.~Pan, Y.~Wu, K.~Cui, \emph{et~al.}, \enquote{Accurate detection and instance segmentation of unstained living adherent cells in differential interference contrast images,} {\protect\JournalTitle{Computers in Biology and Medicine}} \textbf{182}, 109151 (2024).

\bibitem{obara2013bacterial}
B.~Obara, M.~A. Roberts, J.~P. Armitage, and V.~Grau, \enquote{Bacterial cell identification in differential interference contrast microscopy images,} {\protect\JournalTitle{BMC bioinformatics}} \textbf{14}, 1--13 (2013).

\bibitem{ogawa2022different}
T.~Ogawa, K.~Ochiai, T.~Iwata, \emph{et~al.}, \enquote{Different cell imaging methods did not significantly improve immune cell image classification performance,} {\protect\JournalTitle{Plos one}} \textbf{17}, e0262397 (2022).

\bibitem{curl2005refractive}
C.~L. Curl, C.~J. Bellair, T.~Harris, \emph{et~al.}, \enquote{Refractive index measurement in viable cells using quantitative phase-amplitude microscopy and confocal microscopy,} {\protect\JournalTitle{Cytometry Part A: The Journal of the International Society for Analytical Cytology}} \textbf{65}, 88--92 (2005).

\bibitem{wang2010quantitative}
Z.~Wang and G.~Popescu, \enquote{Quantitative phase imaging with broadband fields,} {\protect\JournalTitle{Applied Physics Letters}} \textbf{96} (2010).

\bibitem{park2023artificial}
J.~Park, B.~Bai, D.~Ryu, \emph{et~al.}, \enquote{Artificial intelligence-enabled quantitative phase imaging methods for life sciences,} {\protect\JournalTitle{Nature Methods}} \textbf{20}, 1645--1660 (2023).

\bibitem{nguyen2022quantitative}
T.~L. Nguyen, S.~Pradeep, R.~L. Judson-Torres, \emph{et~al.}, \enquote{Quantitative phase imaging: recent advances and expanding potential in biomedicine,} {\protect\JournalTitle{ACS nano}} \textbf{16}, 11516--11544 (2022).

\bibitem{jiang2022automatic}
M.~Jiang, M.~Shao, X.~Yang, \emph{et~al.}, \enquote{Automatic classification of red blood cell morphology based on quantitative phase imaging,} {\protect\JournalTitle{International Journal of Optics}} \textbf{2022}, 1240020 (2022).

\bibitem{lam2020quantitative}
V.~K. Lam, T.~Nguyen, V.~Bui, \emph{et~al.}, \enquote{Quantitative scoring of epithelial and mesenchymal qualities of cancer cells using machine learning and quantitative phase imaging,} {\protect\JournalTitle{Journal of biomedical optics}} \textbf{25}, 026002--026002 (2020).

\bibitem{roitshtain2017quantitative}
D.~Roitshtain, L.~Wolbromsky, E.~Bal, \emph{et~al.}, \enquote{Quantitative phase microscopy spatial signatures of cancer cells,} {\protect\JournalTitle{Cytometry Part A}} \textbf{91}, 482--493 (2017).

\bibitem{ayyappan2020identification}
V.~Ayyappan, A.~Chang, C.~Zhang, \emph{et~al.}, \enquote{Identification and staging of b-cell acute lymphoblastic leukemia using quantitative phase imaging and machine learning,} {\protect\JournalTitle{ACS sensors}} \textbf{5}, 3281--3289 (2020).

\bibitem{wittenzellner2025label}
K.~Wittenzellner, M.~Lengl, S.~R{\"o}hrl, \emph{et~al.}, \enquote{Label-free single cell phenotyping to determine tumor cell heterogeneity in pancreatic cancer in real-time,} {\protect\JournalTitle{JCI insight}}  (2025).

\bibitem{go2018label}
T.~Go, H.~Byeon, and S.~J. Lee, \enquote{Label-free sensor for automatic identification of erythrocytes using digital in-line holographic microscopy and machine learning,} {\protect\JournalTitle{Biosensors and Bioelectronics}} \textbf{103}, 12--18 (2018).

\bibitem{jaferzadeh2023automated}
K.~Jaferzadeh, S.~Son, A.~Rehman, \emph{et~al.}, \enquote{Automated stain-free holographic image-based phenotypic classification of elliptical cancer cells,} {\protect\JournalTitle{Advanced Photonics Research}} \textbf{4}, 2200043 (2023).

\bibitem{kandel2020phase}
M.~E. Kandel, Y.~R. He, Y.~J. Lee, \emph{et~al.}, \enquote{Phase imaging with computational specificity (pics) for measuring dry mass changes in sub-cellular compartments,} {\protect\JournalTitle{Nature communications}} \textbf{11}, 6256 (2020).

\bibitem{zipfel2003nonlinear}
W.~R. Zipfel, R.~M. Williams, and W.~W. Webb, \enquote{Nonlinear magic: multiphoton microscopy in the biosciences,} {\protect\JournalTitle{Nature biotechnology}} \textbf{21}, 1369--1377 (2003).

\bibitem{yu2009two}
Q.~Yu and A.~A. Heikal, \enquote{Two-photon autofluorescence dynamics imaging reveals sensitivity of intracellular nadh concentration and conformation to cell physiology at the single-cell level,} {\protect\JournalTitle{Journal of Photochemistry and Photobiology B: Biology}} \textbf{95}, 46--57 (2009).

\bibitem{shi2025optical}
L.~Shi and J.~Villazon, \enquote{Optical imaging unveiling metabolic dynamics in cells and organisms during aging and diseases,} {\protect\JournalTitle{Med-X}} \textbf{3}, 1--24 (2025).

\bibitem{shi2017label}
L.~Shi, L.~Lu, G.~Harvey, \emph{et~al.}, \enquote{Label-free fluorescence spectroscopy for detecting key biomolecules in brain tissue from a mouse model of alzheimer’s disease,} {\protect\JournalTitle{Scientific reports}} \textbf{7}, 2599 (2017).

\bibitem{shi2018optical}
L.~Shi, C.~Zheng, Y.~Shen, \emph{et~al.}, \enquote{Optical imaging of metabolic dynamics in animals,} {\protect\JournalTitle{Nature communications}} \textbf{9}, 2995 (2018).

\bibitem{kreiss2020label}
L.~Krei{\ss}, O.-M. Thoma, A.~Dilipkumar, \emph{et~al.}, \enquote{Label-free in vivo histopathology of experimental colitis via 3-channel multiphoton endomicroscopy,} {\protect\JournalTitle{Gastroenterology}} \textbf{159}, 832--834 (2020).

\bibitem{kreiss2022label}
L.~Kreiss, O.-M. Thoma, S.~Lemire, \emph{et~al.}, \enquote{Label-free characterization and quantification of mucosal inflammation in common murine colitis models with multiphoton imaging,} {\protect\JournalTitle{Inflammatory Bowel Diseases}} \textbf{28}, 1637--1646 (2022).

\bibitem{schmalzl2022interferon}
A.~Schmalzl, T.~Leupold, L.~Kreiss, \emph{et~al.}, \enquote{Interferon regulatory factor 1 (irf-1) promotes intestinal group 3 innate lymphoid responses during citrobacter rodentium infection,} {\protect\JournalTitle{Nature Communications}} \textbf{13}, 5730 (2022).

\bibitem{sommer2023anti}
K.~Sommer, K.~Heidbreder, L.~Kreiss, \emph{et~al.}, \enquote{Anti-$\beta$7 integrin treatment impedes the recruitment on non-classical monocytes to the gut and delays macrophage-mediated intestinal wound healing,} {\protect\JournalTitle{Clinical and Translational Medicine}} \textbf{13}, e1233 (2023).

\bibitem{habibalahi2025multispectral}
A.~Habibalahi, A.~G. Anwer, A.~Knab, \emph{et~al.}, \enquote{Multispectral autofluorescence for label free classification of immune cell type and activation/polarization status,} {\protect\JournalTitle{Scandinavian Journal of Immunology}} \textbf{101}, e70004 (2025).

\bibitem{shah2023autofluorescence}
V.~S. Shah, J.~Hou, V.~Vinarsky, \emph{et~al.}, \enquote{Autofluorescence imaging permits label-free cell type assignment and reveals the dynamic formation of airway secretory cell associated antigen passages (saps),} {\protect\JournalTitle{Elife}} \textbf{12}, e84375 (2023).

\bibitem{Lemire2022}
S.~Lemire, O.-M. Thoma, L.~Kreiss, \emph{et~al.}, \enquote{Natural nadh and fad autofluorescence as label-free biomarkers for discriminating subtypes and functional states of immune cells,} {\protect\JournalTitle{International Journal of Molecular Sciences}} \textbf{23} (2022).

\bibitem{abir2024metabolic}
A.~H. Abir, L.~Weckwerth, A.~Wilhelm, \emph{et~al.}, \enquote{Metabolic profiling of single cells by exploiting nadh and fad fluorescence via flow cytometry,} {\protect\JournalTitle{Molecular Metabolism}} \textbf{87}, 101981 (2024).

\bibitem{zhang2020digital}
Y.~Zhang, K.~de~Haan, Y.~Rivenson, \emph{et~al.}, \enquote{Digital synthesis of histological stains using micro-structured and multiplexed virtual staining of label-free tissue,} {\protect\JournalTitle{Light: Science \& Applications}} \textbf{9}, 78 (2020).

\bibitem{Rivenson_VS}
Y.~Rivenson, H.~D. Wang, Z.~S. Wei, \emph{et~al.}, \enquote{Virtual histological staining of unlabelled tissue-autofluorescence images via deep learning,} {\protect\JournalTitle{Nature Biomedical Engineering}} \textbf{3}, 466--477 (2019).

\bibitem{kreiss2023digital}
L.~Kreiss, S.~Jiang, X.~Li, \emph{et~al.}, \enquote{Digital staining in optical microscopy using deep learning-a review,} {\protect\JournalTitle{PhotoniX}} \textbf{4}, 34 (2023).

\bibitem{gehlsen2015non}
U.~Gehlsen, M.~Szaszak, A.~Gebert, \emph{et~al.}, \enquote{Non-invasive multi-dimensional two-photon microscopy enables optical fingerprinting (tpof) of immune cells,} {\protect\JournalTitle{Journal of Biophotonics}} \textbf{8}, 466--479 (2015).

\bibitem{bocklitz2016pseudo}
T.~W. Bocklitz, F.~S. Salah, N.~Vogler, \emph{et~al.}, \enquote{Pseudo-he images derived from cars/tpef/shg multimodal imaging in combination with raman-spectroscopy as a pathological screening tool,} {\protect\JournalTitle{BMC cancer}} \textbf{16}, 1--11 (2016).

\bibitem{pradhan2021computational}
P.~Pradhan, T.~Meyer, M.~Vieth, \emph{et~al.}, \enquote{Computational tissue staining of non-linear multimodal imaging using supervised and unsupervised deep learning,} {\protect\JournalTitle{Biomedical Optics Express}} \textbf{12}, 2280--2298 (2021).

\bibitem{myaing2006fiber}
M.~T. Myaing, D.~J. MacDonald, and X.~Li, \enquote{Fiber-optic scanning two-photon fluorescence endoscope,} {\protect\JournalTitle{Optics letters}} \textbf{31}, 1076--1078 (2006).

\bibitem{bao2008fast}
H.~Bao, J.~Allen, R.~Pattie, \emph{et~al.}, \enquote{Fast handheld two-photon fluorescence microendoscope with a 475 $\mu$m$\times$ 475 $\mu$m field of view for in vivo imaging,} {\protect\JournalTitle{Optics letters}} \textbf{33}, 1333--1335 (2008).

\bibitem{llewellyn2008minimally}
M.~E. Llewellyn, R.~P. Barretto, S.~L. Delp, and M.~J. Schnitzer, \enquote{Minimally invasive high-speed imaging of sarcomere contractile dynamics in mice and humans,} {\protect\JournalTitle{Nature}} \textbf{454}, 784--788 (2008).

\bibitem{rivera2011compact}
D.~R. Rivera, C.~M. Brown, D.~G. Ouzounov, \emph{et~al.}, \enquote{Compact and flexible raster scanning multiphoton endoscope capable of imaging unstained tissue,} {\protect\JournalTitle{Proceedings of the National Academy of Sciences}} \textbf{108}, 17598--17603 (2011).

\bibitem{huland2012vivo}
D.~M. Huland, C.~M. Brown, S.~S. Howard, \emph{et~al.}, \enquote{In vivo imaging of unstained tissues using long gradient index lens multiphoton endoscopic systems,} {\protect\JournalTitle{Biomedical optics express}} \textbf{3}, 1077--1085 (2012).

\bibitem{dilipkumar2019label}
A.~Dilipkumar, A.~Al-Shemmary, L.~Krei{\ss}, \emph{et~al.}, \enquote{Label-free multiphoton endomicroscopy for minimally invasive in vivo imaging,} {\protect\JournalTitle{Advanced Science}} \textbf{6}, 1801735 (2019).

\bibitem{dodt1990visualizing}
H.-U. Dodt and W.~Zieglg{\"a}nsberger, \enquote{Visualizing unstained neurons in living brain slices by infrared dic-videomicroscopy,} {\protect\JournalTitle{Brain research}} \textbf{537}, 333--336 (1990).

\bibitem{lowe2004distinctive}
D.~G. Lowe, \enquote{Distinctive image features from scale-invariant keypoints,} {\protect\JournalTitle{International journal of computer vision}} \textbf{60}, 91--110 (2004).

\bibitem{hastie2009elements}
T.~Hastie, R.~Tibshirani, J.~H. Friedman, and J.~H. Friedman, \emph{The elements of statistical learning: data mining, inference, and prediction}, vol.~2 (Springer, 2009).

\bibitem{iandola2016exploring}
F.~Iandola, \emph{Exploring the design space of deep convolutional neural networks at large scale} (University of California, Berkeley, 2016).

\bibitem{kingma2017adammethodstochasticoptimization}
D.~P. Kingma and J.~Ba, \enquote{Adam: A method for stochastic optimization,}  (2017).

\bibitem{izmailov2019averagingweightsleadswider}
P.~Izmailov, D.~Podoprikhin, T.~Garipov, \emph{et~al.}, \enquote{Averaging weights leads to wider optima and better generalization,}  (2019).

\bibitem{cooke2022multiple}
C.~L. Cooke, K.~Kim, S.~Xu, \emph{et~al.}, \enquote{A multiple instance learning approach for detecting covid-19 in peripheral blood smears,} {\protect\JournalTitle{PLOS Digital Health}} \textbf{1}, e0000078 (2022).

\bibitem{Somani2022}
A.~Somani, A.~A. Sekh, I.~S. Opstad, \emph{et~al.}, \enquote{Virtual labeling of mitochondria in living cells using correlative imaging and physics-guided deep learning,} {\protect\JournalTitle{Biomedical Optics Express}} \textbf{13}, 5495--5516 (2022).

\bibitem{NAEIM201829}
F.~Naeim, P.~{Nagesh Rao}, S.~X. Song, and R.~T. Phan, \enquote{Chapter 2 - principles of immunophenotyping,} in \emph{Atlas of Hematopathology (Second Edition),}  F.~Naeim, P.~{Nagesh Rao}, S.~X. Song, and R.~T. Phan, eds. (Academic Press, 2018), pp. 29--56, second edition ed.

\bibitem{haber2021generation}
L.~Haber, K.~Olson, M.~P. Kelly, \emph{et~al.}, \enquote{Generation of t-cell-redirecting bispecific antibodies with differentiated profiles of cytokine release and biodistribution by cd3 affinity tuning,} {\protect\JournalTitle{Scientific Reports}} \textbf{11}, 14397 (2021).

\bibitem{buckchash2025applications}
H.~Buckchash, G.~K. Verma, and D.~K. Prasad, \enquote{Applications and challenges of ai and microscopy in life science research: A review,} {\protect\JournalTitle{arXiv preprint arXiv:2501.13135}}  (2025).

\end{thebibliography}

\newpage

\section*{Supplemental material}

\begin{figure}
\centering
\includegraphics[width=0.8\linewidth]{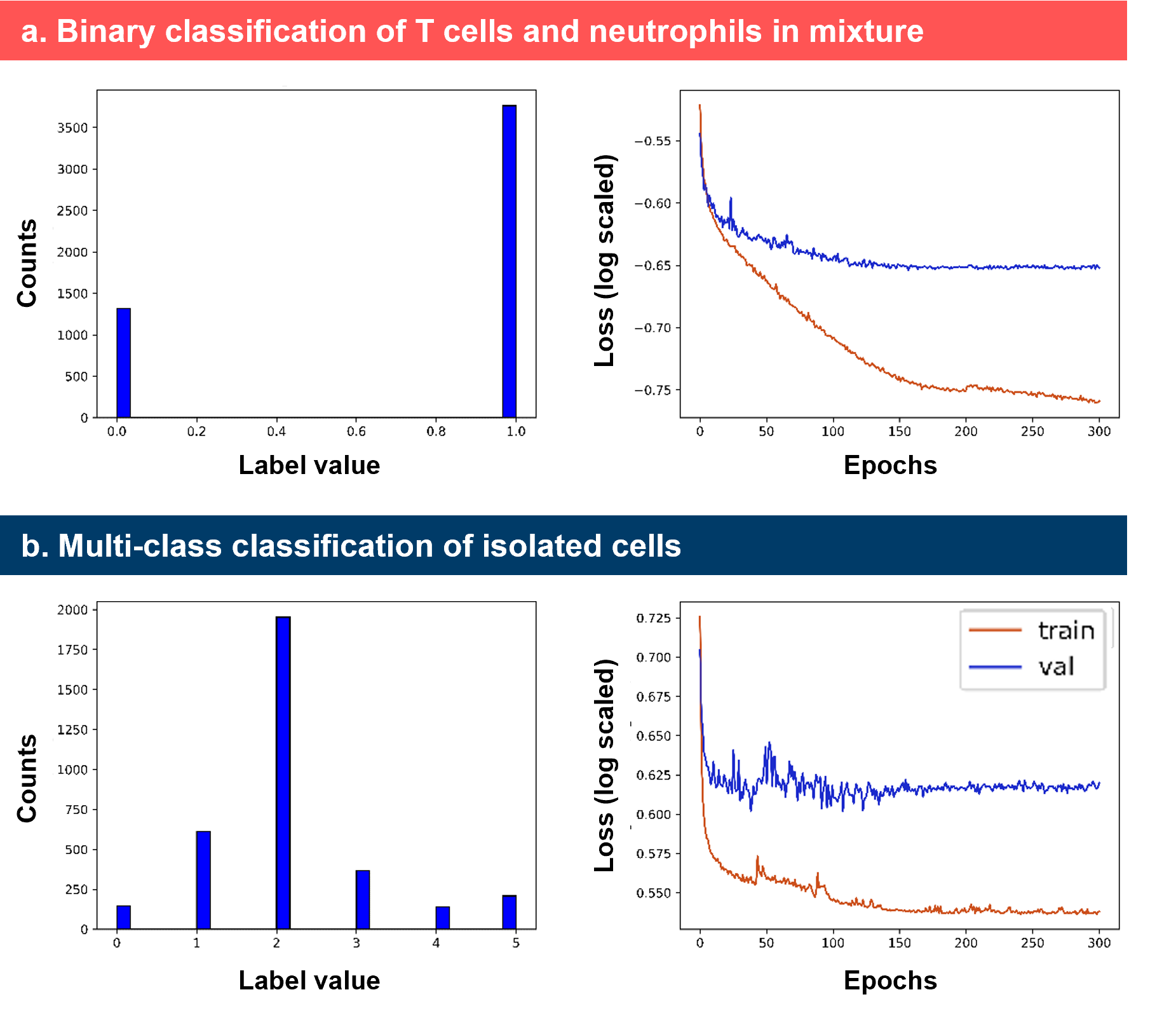}
\caption{Histograms of class labels and learning curves from binary classification of cells in mixture (a) and from multi-class classification from isolated cells (b).}
\label{Supp_Fig_1}
\end{figure}

\begin{table}
    \centering
    \begin{tabular}{lcccccr}
        Block & Layer & Output  & Filter & Stride & Depth & Trainable \\
        & name & size & size  &  &  & parameters \\
        \hline
         0 & Input & 64x64x3 & & & & \\
         1 & Features & 32x32x96 & 7x7 & 2 & 1 & 14,208 \\
         2 & MaxPool & 16x16x96 & 3x3 & 2 & 0 & \\
         3 & Fire & 16x16x128 & & & 2 & 11,920 \\
         4 & Fire & 16x16x128 & & & 2 & 12,432 \\
         5 & Fire & 16x16x256 & & & 2 & 45,344 \\
         6 & MaxPool & 8x8x256 & 3x3 & 2 & 0 & \\
         7 & Fire & 8x8x256 & & & 2 & 49,440 \\
         8 & Fire & 8x8x384 & & & 2 & 104,880 \\
         9 & Fire & 8x8x384 & & & 2 & 111,024 \\
         10 & Fire & 8x8x512 & & & 2 & 188,992 \\
         11 & MaxPool & 4x4x512 & 3x3 & 2 & 0 & \\
         10 & Fire & 4x4x512 & & & 2 & 197,184 \\
         11 & Classifier* & 1x1x1 & & & 2 & 512 \\  
         \hline
          & & & & & & \textbf{735,936} \\
    \end{tabular}
    \caption{Model structure and number of trainable parameters (representation based on table 5.1 in Ref.~\cite{iandola2016exploring} and adjusted for our data). *) The classifier consisted of a dropout layer, a 2D conv layer, a sigmoid and an adaptive average pooling layer.}
    \label{tab_model}
\end{table}

\begin{table}[!htb]
    \caption{Hyperparameters}
      \small
      \centering
        \begin{tabular}{lrr}
        \hline
        \textbf{Parameter} & \textbf{Binary classification} &  \textbf{Multi-class classification} \\
                           & \textbf{Mixture of cells}      & \textbf{Isolated cell types} \\
                           & \textbf{GT from antibody marker}      & \textbf{GT from experimental protocol} \\
        
        \hline
         $N_{classes}$      & 2                         & 6 \\
         $N_{epochs}$       & 300                       & 300 \\
         Learning rate      & 5e-06                     & 5e-06 \\
         Augmentation probability & 0.6                 & 0.0 \\
         Batch size         & 16                        & 32 \\
         Dropout            & 0.1                       & 0.1 \\
         Label smoothing    & 0.0                       & 0.2 \\
         Weight decay       & 0.001                     & 0.0005 \\
    \end{tabular}
    
    \label{tab_hyperparam}
\end{table}

\end{document}